\documentclass[11pt]{article}

\usepackage{naacl2021}

\usepackage{times}
\usepackage{latexsym}

\usepackage[T1]{fontenc}

\usepackage[utf8]{inputenc}

\usepackage{microtype}

\usepackage{amsmath}
\usepackage{bm}

\usepackage{color}
\usepackage{multirow}
\usepackage{amsmath}
\usepackage{subfig}
\usepackage{enumitem}
\usepackage{amsfonts}
\usepackage{multirow}
\usepackage{makecell}
\usepackage{arydshln}
\usepackage{graphicx} 
\usepackage{graphics}
\usepackage{bbm}
\usepackage{balance}
\usepackage{booktabs}
\usepackage{threeparttable}

\usepackage[table]{xcolor}
\usepackage{adjustbox}
\usepackage{booktabs}

\definecolor{jiao}{RGB}{168, 0, 128}
\newcommand{\jiao}[1]{\textcolor{black}{#1}}
\definecolor{ychao}{RGB}{185,85 ,32}

\definecolor{slhe}{RGB}{240, 0, 0}

\definecolor{g1}{RGB}{240,240,255}
\definecolor{g2}{RGB}{210,210,255}
\definecolor{g3}{RGB}{180,180,255}
\definecolor{g4}{RGB}{150,150,255}

\definecolor{r1}{RGB}{255,240,240}
\definecolor{r2}{RGB}{255,210,210}
\definecolor{r3}{RGB}{255,180,180}
\definecolor{r4}{RGB}{255,150,150}

\title{Multi-Task Learning with Shared Encoder for Non-Autoregressive Machine Translation}

\author{
Yongchang Hao$^{\dagger}$\thanks{\ \ The first two authors contributed equally to this work. This work was conducted when Yongchang Hao, Shilin He, and Wenxiang Jiao  were interning at Tencent AI Lab.}  ~~ Shilin He$^{\ddagger\ast}$ ~~ Wenxiang Jiao$^{\ddagger}$   \AND\vspace{0.1in}  ~~ Zhaopeng Tu$^{\dagger\dagger}$  ~~ Michael R. Lyu$^{\ddagger}$  ~~  Xing Wang$^{\dagger\dagger}$ \\
{$^\dagger$School of Computer Science and Technology, Soochow University} \\
{$^\ddagger$Department of Computer Science and Engineering, The Chinese University of Hong Kong} \\
{$^{\dagger\dagger}$Tencent AI Lab} \\
$^\dagger${\tt yongchanghao.w@gmai.com} \\
$^\ddagger${\tt \{slhe,wxjiao,lyu\}@cse.cuhk.edu.hk} \\
$^{\dagger\dagger}${\tt \{zptu,brightxwang\}@tencent.com}  }

\date{}

\begin{document}

\maketitle

\vspace*{0.4in}
\begin{abstract}
Non-Autoregressive machine Translation (NAT) models have demonstrated significant inference speedup but suffer from inferior translation accuracy. The common practice to tackle the problem is transferring the Autoregressive machine Translation (AT) knowledge to NAT models, e.g., with knowledge distillation. In this work, we hypothesize and empirically verify that AT and NAT encoders capture different linguistic properties of source sentences. Therefore, we propose to adopt multi-task learning to transfer the AT knowledge to NAT models through encoder sharing. Specifically, we take the AT model as an auxiliary task to enhance NAT model performance. Experimental results on WMT14  English$\Leftrightarrow$German and  WMT16 English$\Leftrightarrow$Romanian datasets show that the proposed \textsc{multi-task NAT} achieves significant improvements over the baseline NAT models. Furthermore, the performance on large-scale WMT19 and WMT20 English$\Leftrightarrow$German datasets confirm the consistency of our proposed method.  In addition, experimental results demonstrate that our \textsc{multi-task NAT} is complementary to knowledge distillation, the standard knowledge transfer method for NAT. \footnote{\ Code is publicly available at \url{https://github.com/yongchanghao/multi-task-nat}}
\end{abstract}

\section{Introduction}
Neural machine translation (NMT), as the state-of-the-art machine translation paradigm, has recently been approached with two different sequence decoding strategies. The first type \textit{autoregressive \,\,\,\,\,\,\,\,\,\,\phantom. }

\vspace*{0.3in}
\noindent\textit{translation} (AT) models generate output tokens one by one following the left to right direction~\cite{Vaswani:2017:NIPS, bahdanau2014neural}, but it is often criticized for its slow inference speed~\cite{gu2017non}. The second type \textit{non-autoregressive translation} (NAT) models adopt a parallel decoding algorithm to produce output tokens simultaneously~\cite{gu2019levenshtein, ghazvininejad:2019:EMNLP, Ma2020FPETSFP}, but the translation quality of it is often inferior to auto-regressive models~\cite{gu2017non}.

Many researchers have investigated the collaboration between AT and NAT models.
For instance, \textsc{Encoder-NAD-AD}~\cite{zhou2020improving} leverages NAT models to improve the performance of AT. Specifically, their method inserts a NAT decoder between the conventional AT encoder and decoder to generate coarse target sequences for the final autoregressive decoding.
A line of research~\cite{wang2019non,guo2019fine,ding2020context} holds the opinion that the lack of contextual dependency on target sentences potentially leads to the deteriorated performance of NAT models. To boost the NAT translation performance, many recent works resort to the knowledge transfer from a well-trained AT model. Typical knowledge transfer methods include sequence-level knowledge distillation with translation outputs generated by strong AT models~\cite{gu2019levenshtein,ghazvininejad:2019:EMNLP}, word-level knowledge distillation with AT decoder representations~\cite{wei2019:acl,li2019hint}, and fine-tuning on AT model by curriculum learning~\cite{guo2019fine}, etc.

In this work, we first verify our our hypothesis that AT and NAT encoders -- although they belong to the same sequence-to-sequence learning task -- capture different linguistic properties of source sentences. We conduct our verification by evaluating the encoder on a set of probing tasks~\cite{conneau2018probing,W18-5431} for AT and NAT models. Further, by leveraging the linguistic differences, we then adopt a multi-task learning framework with a shared encoder (i.e., \textsc{Multi-Task NAT}) to transfer the AT model knowledge into the NAT model. 

Specifically, we employ an additional AT task as the auxiliary task of which the encoder parameters are shared with the NAT task while parameters of the decoder are exclusive. Since many works~\cite{cipolla2018multitask, Liu_2019_CVPR} suggest that the weights for each task are critical to the multi-task learning, in this work, the multi-task weight assigned to the AT task is dynamically annealed from $1$ to $0$. We name this scheme \textit{importance annealing}. We empirically show the benefit of \textit{importance annealing} in both directions of the original WMT14 English$\Leftrightarrow$German dataset.

Further with knowledge distillation, our proposed \textsc{Multi-Task NAT} achieves significant improvements on WMT14 English$\Leftrightarrow$German and WMT16 English$\Leftrightarrow$Romanian datasets. This confirms the effectiveness of our proposed model on machine translation tasks.

Our contributions are as follows:
\begin{itemize}[itemsep=1pt,topsep=0pt,parsep=0pt,partopsep=0pt]
    \item We propose a multi-task learning framework to boost NAT translation quality by transferring the AT knowledge to the NAT model.
    \item Our analyses reveal that the encoder sharing is necessary for capturing more linguistic and semantic information.
    \item Experiments on standard benchmark datasets demonstrate the effectiveness of the proposed \textsc{Multi-Task NAT}. 
\end{itemize}

\begin{table}[t]
  \centering
\begin{tabular}{ l c  c   c  c }
\toprule
 \multicolumn{2}{c}{\bf Task}    &   {\bf AT} &  {\bf NAT} \\
 \midrule
  \multirow{2}{*}{\em Surface }   & SeLen & 91.7            & \textbf{93.4}  \\
                                  & WC    & 76.0            & \textbf{79.1}  \\
 \midrule
  \multirow{3}{*}{\em Syntactic } & TrDep & 45.8            & 46.0  \\  
                                  & ToCo  & 78.3            & \textbf{79.7}  \\
                                  & BShif & \textbf{74.8}            & 73.4   \\   
 \midrule
 \multirow{5}{*}{\em Semantic}    & Tense & 89.2            & 89.2   \\   
                                  & SubN  & 86.2            & \textbf{87.5}   \\   
                                  & ObjN  & 85.2            & 85.3   \\   
                                  & SoMo  &\textbf{54.0}    & 53.0   \\   
                                  & CoIn  & \textbf{64.9}   & 62.8   \\   
\bottomrule
\end{tabular}
  \caption{Performance on the probing tasks of evaluating linguistic properties embedded in the learned representations of AT and NAT models. }
  \label{tab:probing}
\end{table}

\section{Why Shared Encoder?}
\label{sec:shared_encoder}
To verify our hypothesis that AT and NAT encoders capture different linguistic properties of source sentences and can thereby complement each other, we probe the linguistic knowledge~\cite{conneau2018probing} that embedded in the AT and NAT encoders on a set of tasks to investigate to what extent an encoder captures the linguistic properties. We present the  detail for each probing tasks in Appendix~\ref{sec:apdx:prob}. Moreover, \jiao{in Appendix~\ref{sec:apdx:cca}, we also provide a qualitative investigation to capture the difference between high-dimensional representations of AT and NAT encoders from another perspective.}

The AT and NAT models referred to in the following experiments are \textsc{Transformer} and \textsc{Mask-Predict}. We train the models on the WMT14 English$\Rightarrow$German dataset, and the details of the experiments are introduced in the Appendix.

\paragraph{Probing Tasks}

Probing tasks~\cite{conneau2018probing}  can quantitatively measure the linguistic knowledge embedded in the model representation. We follow \newcite{wang2019self} to set model configurations.  The experimental results are depicted in Table~\ref{tab:probing}.

Table~\ref{tab:probing} shows the AT and NAT encoders capture different linguistic properties of source sentences. We observe that on average, the NAT model captures more surface features but less semantic features than the AT model. For example, on the sentence length prediction (SeLen) task, NAT models significantly outperform AT models since the sentence length prediction is a key component in NAT models.

However, for sentence modification (SoMo) and coordinate clauses invertion (CoIn) tasks, the AT model outperforms the NAT model by a large margin. The linguistic probing results reveal that AT and NAT models capture different linguistic properties, which thereby leaves space for the encoder sharing structure.

\section{Approach}\label{sec:appr}

\begin{figure}[!t]
    \centering
    \includegraphics[width=\linewidth]{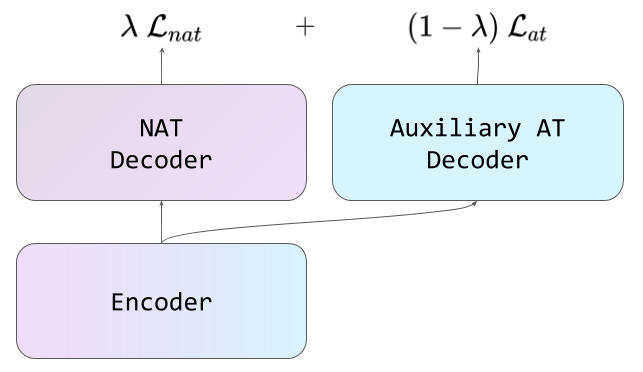}
    \caption{The architecture of our proposed model. We introduce an extra AT decoder to boost the training at the beginning, and gradually lower the importance weight of AT task by increasing $\lambda$.}
    \label{fig:model}
\end{figure}

In this section, we introduce that our shared encoder structure between AT and NAT models under the multi-task learning framework.

\paragraph{Multi-Task NAT}  
Given the AT and NAT models under the standard encoder-decoder structure, we employ the hard parameter sharing method~\cite{ruder2017overview} to share their encoder parameters.  

Therefore, \jiao{as shown in Figure~\ref{fig:model}}, the proposed model \textsc{Multi-Task NAT} consists of three components: shared encoder, AT decoder, and NAT decoder. Their parameters are jointly optimized towards minimizing the multi-task loss function, as introduced in the next section.

\paragraph{Multi-Task Framework}
The loss function  of the proposed  \textsc{Multi-Task NAT} $\mathcal{L}$  at iteration step $t$  is defined as the weighted sum of AT loss and NAT loss:

\begin{equation}
    \begin{aligned}
         \mathcal{L} =  & \lambda_t \mathcal{L}_\mathrm{nat}\left(X, Y; \theta_\mathrm{enc}, \theta^\mathrm{nat}_\mathrm{dec}\right)   \\
     & + (1- \lambda_t) \mathcal{L}_\mathrm{at}\left(X, Y; \theta_\mathrm{enc},\theta^\mathrm{at}_\mathrm{dec}\right)
    \end{aligned}
\label{eq_loss}
\end{equation}
where $\mathcal{L}_\mathrm{at}$ and $\mathcal{L}_\mathrm{nat}$ are AT loss and NAT loss. $\theta_\mathrm{enc}, \theta^\mathrm{nat}_\mathrm{dec}$, and $\theta^\mathrm{at}_\mathrm{dec}$ are parameters of the shared encoder, NAT decoder, and AT decoder respectively. $\lambda_t$ is the importance factor to balance the preference between the AT and NAT models at time step $t$ as illustrated bellow.

\paragraph{Importance Annealing}

The term $\mathcal{L}_\mathrm{at}$ only serves as an auxiliary and does not directly affect the inference of NAT. Therefore, we intuitively determine to lower the importance of the AT loss when the training process is close to the ending, which we named \textit{importance annealing}. Formally, we set $$\lambda_t = \frac{t}{T}$$
where $T$ is the total steps of training. Under such a scheme, the weight for $\mathcal{L}_\mathrm{at}$ is linearly annealed from $1.0$ to $0.0$ along the training process, while the weight for $\mathcal{L}_\mathrm{nat}$ is increased from $0.0$ to $1.0$.

\paragraph{Training and Inference}
During the model training with training pairs $(X, Y)$, we feed the source sentence $X$ to the encoder and the target sentence $Y$ to two decoders separately. The target sentence $Y$ can be either the target sentence in the raw training data (\ref{sec:ablation}) or the generated target sentence with knowledge distillation (\ref{sec:main_result}). During the model inference, we only use the NAT decoder to generate the target tokens simultaneously while ignoring the AT decoder. Therefore, the inference overhead is the same as the NAT model before sharing.

\begin{table}[t]
\centering
\begin{threeparttable}
    \begin{tabular}{l c c}
    \toprule
    \multirow{2}{*}{\bf Model} &   \multicolumn{2}{c}{\bf WMT14} \\
    \cmidrule{2-3} 
    & {\bf En$\Rightarrow$De} &  {\bf De$\Rightarrow$En}\\
    \midrule
    \textsc{Mask-Predict}\tnote{1} &  24.61  &  --  \\
    \textsc{Transformer-Lev}\tnote{2} &  25.20  &  --  \\
    \hdashline
    \textsc{Mask-Predict}\tnote{3} &  24.70  &  29.52  \\
    \hline
    \textsc{Multi-Task NAT}    &    25.66  &  30.09 \\   
    ~~~+ \textit{Importance Annealing}  &    25.79   &  30.32  \\   
    \bottomrule
    \end{tabular}
  \begin{tablenotes}\footnotesize
  \item $^1$~\newcite{ghazvininejad:2019:EMNLP}; $^2$~\newcite{gu2019levenshtein};
  \item $^3$~Our implementation.
  \end{tablenotes}
  \end{threeparttable}
  \caption{Evaluation of translation performance on WMT14 En$\Rightarrow$De and WMT14 De$\Rightarrow$En  test sets without knowledge distillation.}
  \label{table:exp}
\end{table}
\begin{table*}[t]
\centering
    \begin{tabular}{l l l l l}
    \toprule
    \multirow{2}{*}{\bf Model} & \multicolumn{2}{c}{\bf WMT14}  & \multicolumn{2}{c}{\bf WMT16}  \\
    & {\bf En$\Rightarrow$De} &  {\bf De$\Rightarrow$En} & {\bf En$\Rightarrow$Ro} & {\bf Ro$\Rightarrow$En}  \\
    \midrule
   \multicolumn{5}{c}{{\em  Baseline Models}} \\
    Transformer~\cite{ghazvininejad:2019:EMNLP}        & 27.74 & 31.09 & 34.28 & 33.99 \\
    \hdashline
    Hint-based NAT~\cite{li2019hint}                   & 25.20 & 29.52 &  --  & -- \\
    NAT-REG~\cite{wang2019non}           &24.61  & 28.90 &  --  & -- \\
    FCL-NAT~\cite{guo2019fine}                         & 25.75 & 29.50 &  --  & -- \\
    Levenshtein Transformer~\cite{gu2019levenshtein}   & 27.27 & --     & --   & 33.26  \\
    Mask-Predict~\cite{ghazvininejad:2019:EMNLP}       & 27.03 & 30.53 & 33.08 & 33.31 \\
    Mask-Predict w/ Raw Data Prior~\cite{ding2021understanding} & 27.8 & -- & -- &  33.7 \\
    \midrule
    \multicolumn{5}{c}{{\em Our Experiments}} \\
    Mask-Predict            & 27.18 & 30.86 & 33.03 & 32.71\\ 
    \textsc{Multi-Task NAT} (w/ IA) & \bf{27.98}$^\Uparrow$ & \bf{31.27}$^\uparrow$ & \bf{33.80}$^\uparrow$ & \textbf{33.60}$^\Uparrow$  \\
    \bottomrule
    \end{tabular}
  \caption{Evaluation of translation performance on WMT14 En$\Leftrightarrow$De and WMT16 En$\Leftrightarrow$Ro test sets. All NAT models are trained with AT knowledge distillation. ``$\uparrow/\Uparrow$'':  indicate a statistically significant improvement over the corresponding baseline $p < 0.05/0.01$ respectively. Some baselines do not perform well because they are not built upon the \textsc{Mask-Predict}. } 
  \label{table:exp2}
\end{table*}
\section{Experiment}

We conducted experiments on two widely used WMT14 English$\Leftrightarrow$German and WMT16 English$\Leftrightarrow$Romanian benchmark datasets, which consist of 4.5M and 610K sentence pairs, respectively. We applied BPE~\cite{Sennrich:BPE} with 32K merge operations for both language pairs. The experimental results are evaluated in case-sensitive BLEU score~\cite{papineni2002bleu}.

We use \textsc{Transformer}~\cite{Vaswani:2017:NIPS} as our baseline autoregressive translation model and the \textsc{Mask-Predict}~\cite{ghazvininejad:2019:EMNLP} as our baseline non-autoregressive model. We integrate the \textsc{Transformer} decoder into the \textsc{Mask-Predict} to implement the proposed \textsc{Multi-Task NAT} model. For $\lambda_t$, we use the annealing scheme described in Section~\ref{sec:appr}. Since the major NAT architecture of our method is exactly the \textsc{Mask-Predict} model, any established decoding latency results~\cite{kasai2021deep} for \textsc{Mask-Predict} can also be applied to ours. All of the parameters are randomly initialized for a fair comparison with the \textsc{Mask-Predict}. More training details are introduced in Appendix~\ref{sec:apdx:impl}.
\subsection{Ablation Study}\label{sec:ablation}

Table~\ref{table:exp} shows that the performance of our \textsc{Multi-Task NAT} model and baseline models on WMT14 En$\Leftrightarrow$De datasets without using the knowledge distillation. The vanilla \textsc{Multi-Task NAT} model with the
the $\lambda$ fixed as $0.5$ outperforms the baseline \textsc{Mask-Predict} model by 0.96 and 0.57 BLEU score in En$\Rightarrow$De and De$\Rightarrow$En direction respectively and even surpasses the strong baseline \textsc{Transformer-Lev} by 0.46 BLEU points in En$\Rightarrow$De translation. With the \textit{importance annealing}, the \textsc{Multi-Task NAT} model achieves slight but consistent improvements over the vanilla model (``+\textit{Importance Annealing}'' in Table~\ref{table:exp}). The improvements demonstrate the effectiveness of our proposed model using multi-task learning.

\subsection{Main Result}
\label{sec:main_result}
We further evaluate the proposed \textsc{Multi-Task NAT} model with the standard practice of knowledge distillation.  Table~\ref{table:exp2} depicts the performances of our model as well as strong baseline models. Our proposed \textsc{Multi-Task NAT} model achieves a significant improvement of 0.80 and 0.41 BLEU point over the strong baseline \textsc{Mask-Predict} model on En$\Rightarrow$De and De$\Rightarrow$En translation. On En$\Leftrightarrow$Ro translation, our model outperforms the baseline model by 0.77 and 0.89 BLEU scores respectively.  We use the \texttt{compare-mt}~\cite{compare-mt}\footnote{https://github.com/neulab/compare-mt} to determine the significance. Details for significance tests are described in Appendix~\ref{sec:apdx:impl:significance}.

\begin{table}[t]
  \centering
\begin{tabular}{ l c  c   r  r }
\toprule
 \multicolumn{2}{c}{\bf Task}    &   {\bf \textsc{Ours}} &  {$\Delta_{AT}$}  & {$\Delta_{NAT}$} \\
 \midrule
  \multirow{2}{*}{\em Surface }   &  SeLen &  94.4 & \cellcolor{g4} 2.7   & \cellcolor{g3} 1.0 \\
                                  & WC    & 79.3 & \cellcolor{g4} 3.3	 & \cellcolor{g1} 0.2 \\
 \midrule
  \multirow{3}{*}{\em Syntactic } & TrDep & 47.2 & \cellcolor{g3}1.4    &\cellcolor{g3}	1.2 \\  
                                  & ToCo  & 79.3 & \cellcolor{g3}1.0    &\cellcolor{r1}	-0.4 \\
                                  & BShif & 74.7 & \cellcolor{r1}-0.1    & \cellcolor{g3}1.3 \\   
 \midrule
 \multirow{5}{*}{\em Semantic}    & Tense & 88.9 & \cellcolor{r1} -0.3	 &  \cellcolor{r1} -0.3 \\   
                                  & SubN  & 87.1 & \cellcolor{g2}0.9	 &  \cellcolor{r1} -0.4 \\   
                                  & ObjN  & 85.8 & \cellcolor{g2}0.6   & \cellcolor{g2} 0.5 \\   
                                  & SoMo  & 54.8 & \cellcolor{g2}0.8	 & \cellcolor{g3} 1.8 \\   
                                  & CoIn  & 63.3 & \cellcolor{r3} -1.6	 & \cellcolor{g2} 0.5 \\   
\bottomrule
\end{tabular}
  \caption{Performance on the probing tasks of our \textsc{Multi-task NAT} model. {$\Delta_{AT}$} and {$\Delta_{NAT}$} denote the relative increase over the AT and NAT probing performance, respectively.}
  \label{tab:probing_multiNAT}
\end{table} 

\begin{table*}[t]
\centering
\begin{threeparttable}
    \begin{tabular}{l l l l l}
    \toprule
    \multirow{2}{*}{\bf Model} &   \multicolumn{2}{c}{\bf WMT19} &   \multicolumn{2}{c}{\bf WMT20}  \\
     & {\bf En$\Rightarrow$De} &  {\bf De$\Rightarrow$En} & {\bf En$\Rightarrow$De} &  {\bf De$\Rightarrow$En}\\
    \midrule
    \textsc{Mask-Predict}     &  34.79  &  37.04  & 25.24 & 36.36 \\
    \textsc{Multi-Task NAT}   &  \textbf{35.38}$^\uparrow$  &  \textbf{37.62}$^\uparrow$ & \textbf{25.72}$^\uparrow$ & \textbf{36.58}   \\ 
    \bottomrule
    \end{tabular}
  \end{threeparttable}
  \caption{Evaluation of translation performance on WMT19 En$\Leftrightarrow$De and WMT20 En$\Leftrightarrow$De test sets with knowledge distillation. The \textit{Importance Annealing} is adopted by default.}
  \label{table:exp_large}
\end{table*}

\subsection{Analysis}
We conduct probing tasks to empirically reconfirm our hypothesis in Section~\ref{sec:shared_encoder} and better understand our \textsc{Multi-Task NAT} in terms of linguistic properties. The results are presented in Table~\ref{tab:probing_multiNAT}. In most of the cases, our \textsc{Multi-Task NAT} could learn better surface, syntactic, and semantic information than the \textsc{Transformer} and \textsc{Mask-Predict} baseline models, indicating that our multi-task learning framework can indeed take the advantages of two separate tasks and capture better linguistic properties. Notably, on the sentence length (Selen) prediction task and tree depth (TrDep) task, the \textsc{Multi-task NAT} shows significantly better performance. On other tasks, our model demonstrates better or on-par performance compared to the NAT model. Regarding the coordination inversion (CoIn) task, though the \textsc{Multi-Task NAT} shows certainly lower performance than  the \textsc{Transformer}, it still outperforms the \textsc{Mask-Predict} by $0.5$.

\subsection{Large-scale Experiments}
We conduct the larger-scale experiments on the WMT English$\Leftrightarrow$German. We adopt newstest2019 and newstest2020 as the test sets.  The parallel data consists of about $36.8$M sentence pairs. We average the last 5 checkpoints as the final model. The results are listed in Table~\ref{table:exp_large}. The improvements suggest that our model are consistently effective on various scale of data.

\section{Conclusion and Future Work}

In this paper, we have presented a novel multi-task learning approach for NAT model with a hard parameter sharing mechanism. Experimental results confirm the significant effect of the proposed \textsc{Multi-Task NAT} model, which shows the complementary effects of multi-task learning to the knowledge distillation method. 

Based on our \textsc{Multi-Task NAT}, there are many promising directions for future research. For example, 1) decoder interaction: knowledge distillation in an online fashion between AT and NAT decoders; 2) share-all framework: shared-encoder and shared-decoder with two decoding strategies, and the model can dynamically choose the optimal decoding strategy during model inference. 3) data manipulation strategies: such as data rejuvenation~\cite{jiao2020data}, lexical frequency discrepancy~\cite{ding2021understanding}.

\section*{Acknowledgements}
The authors sincerely thank Liang Ding for the advice of experiments settings, and the anonymous reviewers for their insightful suggestions on various aspects of this work.

\bibliography{anthology,custom}
\bibliographystyle{acl_natbib}

\clearpage

\appendix

\section{Implementation}
\label{sec:apdx:impl}

We implemented the proposed \textsc{Multi-Task NAT} model based on Fairseq~\cite{ott:2019:naacl}, an open-source framework for the sequence to sequence learning. More specifically, we added the Transformer decoder (AT) to the standard Mask-Predict (NAT) model, and the encoder output is fed into the Transformer decoder with the encoder-decoder attention. We implemented a new loss function by combining the AT loss and NAT loss, and we jointly optimized all parameters. We will make all the code publicly available for future use.

\subsection{Hyperparameters}
For the NAT baseline model, we followed the hyperparameter settings as described in Mask-Predict~\cite{ghazvininejad:2019:EMNLP}. More specifically, we trained all models for up to 300K steps with 128K (16000$\times$8) tokens per batch using Adam~\cite{kingma2014adam} with $\beta = (0.9, 0.98)$ and $\epsilon = 10^{-6}$. We adopted the warm-up learning rate scheduler, which linearly increases from $10^{-7}$ to a peak of $5 \cdot 10^{-4}$ within 10,000 steps, and then decays with the inverse square root schedule. 

For the AT baseline model in our preliminary experiments, to make a fair comparison, we reused most parameter settings (e.g., training steps, initialization method, warm-up schedule) of the NAT model to train a strong AT baseline. Since half of the training tokens are randomly masked in the NAT baseline model, we set the batch size in the AT model as 64K (8000 $\times$ 8) tokens.

For our \textsc{Multi-Task NAT} model, we followed the same parameter setting as the NAT baseline model. We saved checkpoints every 2,000 steps and average the last 5 checkpoints as the final model. Due to the limited training data in the En$\Leftrightarrow$Ro translation task, we adopted the early stopping to prevent over-fitting in both the baseline NAT model and our model. 

In the inference phase, all NAT models and our model were using the iterative decoding strategy to perform non-autoregressive translation, with the max decoding steps of 10 and length beams of 5.


\subsection{Model Training}
We applied the mixed precision training to all models to accelerate the training speed. The baseline model took around 15 hours to finish training on 8 Nvidia V100 GPU, while ours took around 30 hours. During the training phase, our model had an extra decoder of 25M parameters, but the parameter size (around 64M) was the same as the baseline model in inference.

\begin{figure}[t!]
    \centering
    \vspace{0.2in}
    \includegraphics[width=0.8\linewidth]{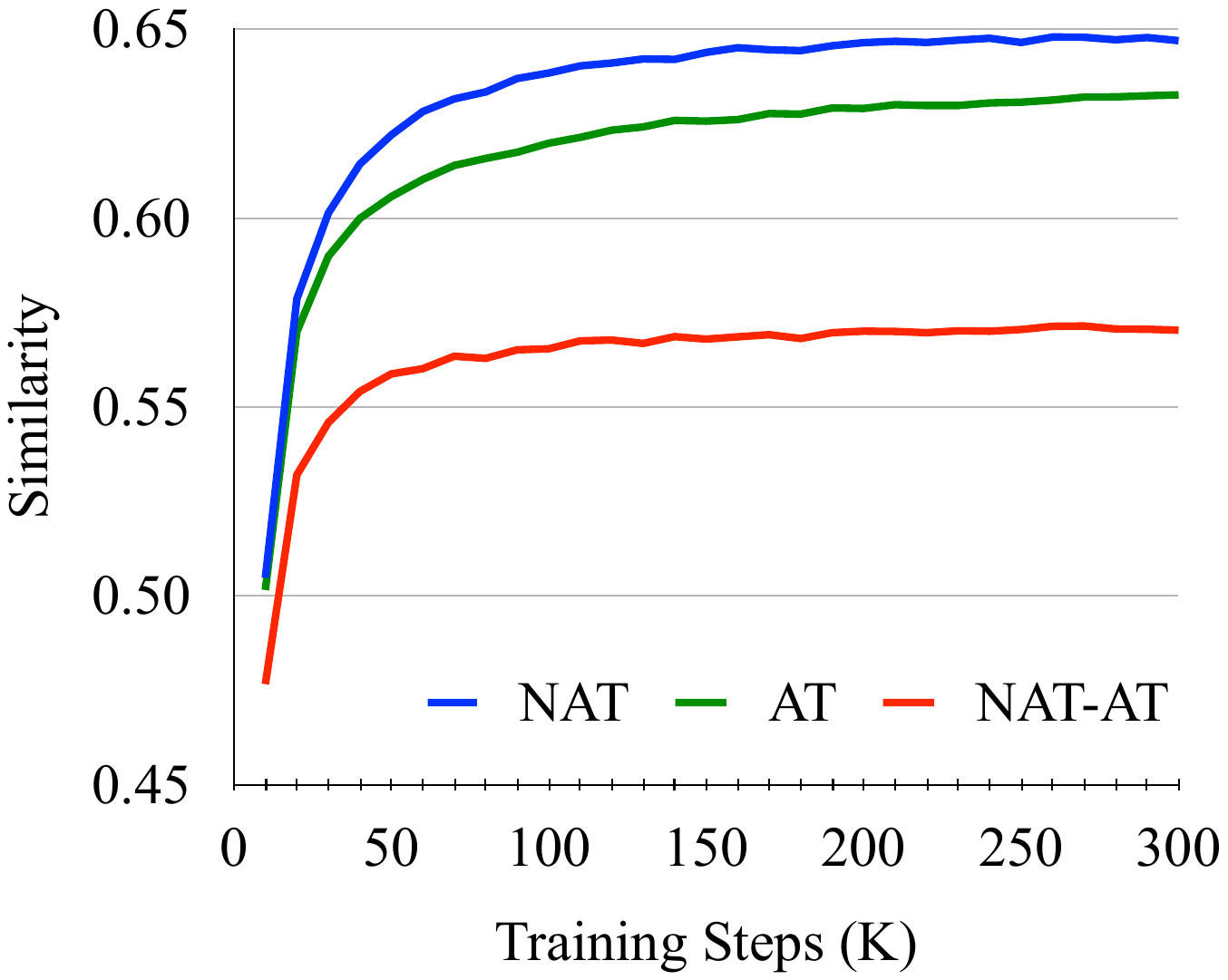} 
    \caption{Representation similarity evolvement during the learning course, NAT (or AT) denotes the baseline of representation similarity between two baselines NAT (or AT) models with different initialization seeds. NAT-AT denotes the representation similarity between NAT and AT encoder. Best viewed in color.}
\label{fig:pwcca}
\end{figure}

\subsection{Model Inference}
In our model, the auxiliary AT decoder is employed to better capture the source sentence representation during the training phase. However, since our focus is on the NAT translation performance, the auxiliary AT decoder is not necessary during the inference. We simply ignored the AT decoder in our approach to keep the same translation speed as the baseline NAT models, leaving the space for future works to leverage the ignored AT decoder and benefit the translation.

\subsection{Significance Tests}
\label{sec:apdx:impl:significance}
We use the \texttt{compare-mt}~\cite{compare-mt} with 1000 re-samples to perform statistical significance tests for the \textsc{Mask-Predict} model output and our model output. The results show that our model significantly outperforms the baseline model in all language directions. All these experimental results confirm the effectiveness of our proposed method even with knowledge distillation. Besides, considering the results in Table~\ref{table:exp}, as a new knowledge transfer approach, our proposed multi-task learning method \textsc{Multi-Task NAT} can complement the classic knowledge distillation method, which is promising for future exploration.

\section{Probing Tasks}
\label{sec:apdx:prob}

We conducted 10 probing tasks\footnote{https://github.com/facebookresearch/SentEval/tree/master\\/data/probing} to study what linguistic properties are captured by the encoder~\cite{conneau2018senteval}.
A probing task is a classification problem that focuses on simple linguistic properties of sentences. `SeLen' predicts the number of words in a sentence. `WC' predicts which of the target words appear on the given sentence given its sentence embedding. `TrDep' checks whether the encoder representation infers the hierarchical structure of sentences. In `ToCo' task, it measures the sequence of top constituents immediately below the sentence node. `BShif' predicts whether two consecutive tokens within the sentence have been inverted or not. `Tense' asks for the tense of the main-clause verb. `SubN' focuses on the number of the main clause's subject. `ObjN' studies the number of the direct object of the main clause. In `SoMo', some sentences are modified by replacing a random noun or verb with another one and the classifier should identify whether a sentence has been modified. `CoIn' contains sentences made of two coordinate clauses. Half of the sentences have inverted the order of the clauses and the task is to tell whether a sentence is intact or modified.

We first extracted the sentence representations of input from the AT and NAT encoder, which were used to carry out our probing tasks. For both the AT model and NAT model, the mean of the encoder top layer representations was used as the sentence representation. The classifier we used in this work was a Multi-Layer Perceptron (MLP) with a hidden dimension of 200. We optimized the model using the Adam optimizer with a learning rate of 0.001. The batch size was set to 64 and we trained the model for 4 epochs. The 10-fold cross-validation was also employed to get the final performance.

\section{Representation Similarity}
\label{sec:apdx:cca}

In this section, we adopt CCA~\cite{morcos2018insights}, a widely-utilized representation comparison metric~\cite{saphra-lopez:2019:NAACL,voita-etal:2019:EMNLP}, to calculate the encoder representation similarity throughout the training course. The CCA score ranges from 0 to 1, and a higher CCA score indicates both encoder representations contain more similar information.

In our experiments, we first extracted the sentence encoder representations for 100,000 training examples (which is sufficient enough for CCA calculation) for each model. Then, we calculated the CCA score between the encoder representations of two models, where the representation was the mean of the top-layer encoder representations. In addition, to demonstrate the evolvement of the representation similarity during the learning course, we calculated the CCA score between two models for every 10,000 training steps.

Figure~\ref{fig:pwcca} shows the evolvement of representation similarity during the model training. We compute the CCA similarity between two NAT (or AT) models under different initialization seeds and take the similarity as the baseline (green and blue lines in Figure~\ref{fig:pwcca}). Comparing to the baseline similarity of NAT (or AT), a lower similarity between AT and NAT representations (red line in Figure~\ref{fig:pwcca}) indicates that AT and NAT encoders capture different source information.

\end{document}